\documentclass[letterpaper]{article} 
\usepackage{aaai25}  
\usepackage{times}  
\usepackage{helvet}  
\usepackage{courier}  
\usepackage[hyphens]{url}  
\usepackage{graphicx} 
\usepackage{natbib}  
\usepackage{caption} 
\usepackage{algorithm}
\usepackage{algorithmic}
\usepackage{amsmath}
\usepackage{multirow}
\usepackage{siunitx}
\usepackage{booktabs}

\usepackage{newfloat}
\usepackage{listings}
\DeclareCaptionStyle{ruled}{labelfont=normalfont,labelsep=colon,strut=off} 
\floatstyle{ruled}
\newfloat{listing}{tb}{lst}{}
\floatname{listing}{Listing}
\title{Evaluating Precise Geolocation Inference Capabilities\\of Vision Language Models}
\author {
    Neel Jay\textsuperscript{\rm 1 \rm 2},
    Hieu Minh ``Jord" Nguyen\textsuperscript{\rm 1 \rm 4},
    Trung Dung Hoang\textsuperscript{\rm 1 \rm 3},
    Jacob Haimes\textsuperscript{\rm 1}
}
\affiliations {
    \textsuperscript{\rm 1}Apart Research\\
    \textsuperscript{\rm 2}University of Maryland\\
    \textsuperscript{\rm 3}Hanoi University of Science and Technology\\
    \textsuperscript{\rm 4}University of Science and Technology of Hanoi\\
    njay@terpmail.umd.edu, jordnguyen43@gmail.com, dung.ht235488@sis.hust.edu.vn, jacob@apartresearch.com
}

\begin{document}

\maketitle

\begin{abstract}
The prevalence of Vision-Language Models (VLMs) raises important questions about privacy in an era where visual information is increasingly available. While foundation VLMs demonstrate broad knowledge and learned capabilities, we specifically investigate their ability to infer geographic location from previously unseen image data. This paper introduces a benchmark dataset collected from Google Street View that represents its global distribution of coverage. Foundation models are evaluated on single-image geolocation inference, with many achieving median distance errors of $<$300 km. We further evaluate VLM ``agents" with access to supplemental tools, observing up to a 30.6\% decrease in distance error. Our findings establish that modern foundation VLMs can act as powerful image geolocation tools, without being specifically trained for this task. When coupled with increasing accessibility of these models, our findings have greater implications for online privacy. We discuss these risks, as well as future work in this area.\footnote{Datasets and code can be found at: \url{https://github.com/njspyx/location-inference}}
\end{abstract}

\begin{table}[t]
\centering
\small
\begin{tabular}{@{}lcr@{}}
\multicolumn{3}{c}{(a) Population Distribution}\\
\\
\toprule
\textbf{Class} & \textbf{Population Range} & \textbf{Count} \\\midrule
Small Urban & 50,000--200,000 & 560 \\
Medium Urban & 200,000--500,000 & 539 \\
Metropolitan & 500,000--1.5M & 323 \\
Large Metropolitan & $>$1.5M & 141 \\
\bottomrule
\end{tabular}

\begin{tabular}{@{}lr@{}}
\\
\multicolumn{2}{c}{(b) Geographic Distribution} \\
\\
\toprule
\textbf{Continent} & \textbf{Count} \\
\midrule
North America & 26 \\
Asia & 21 \\
Europe & 21 \\
Africa & 13 \\
South America & 5 \\
Oceania & 2 \\
\bottomrule
\end{tabular}
\caption{Geographical and population distribution of full benchmark dataset.}
\label{tab:dataset-dist}
\end{table}

\begin{figure*}[t]
\centering
\includegraphics[width=\textwidth]{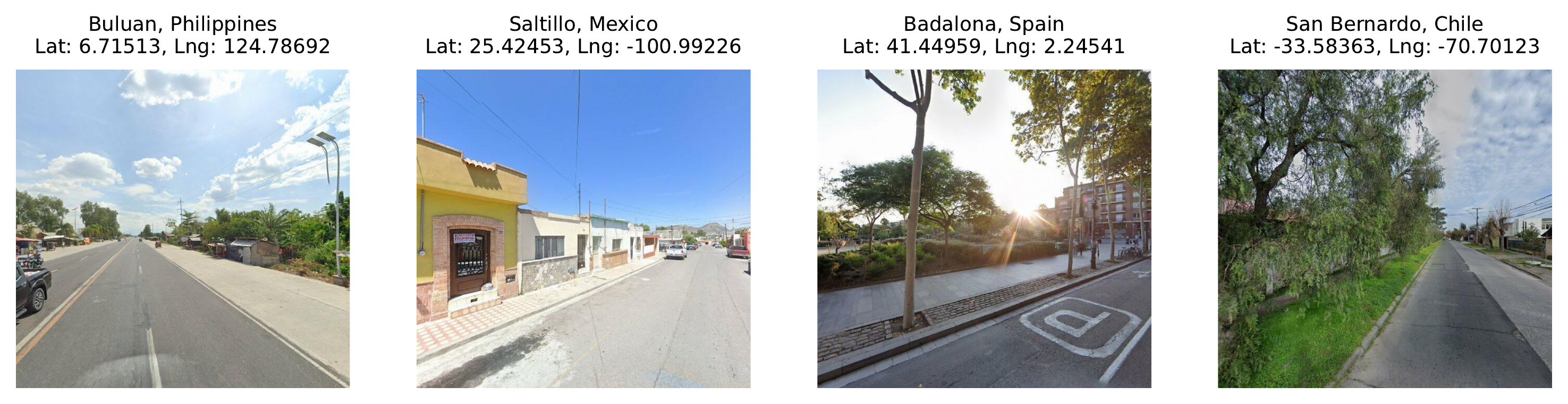} 
\caption{A snapshot of Google Street View images in the full benchmark dataset.}
\label{fig:snapshot}
\end{figure*}

\begin{figure*}[t]
\centering
\includegraphics[scale=0.755]{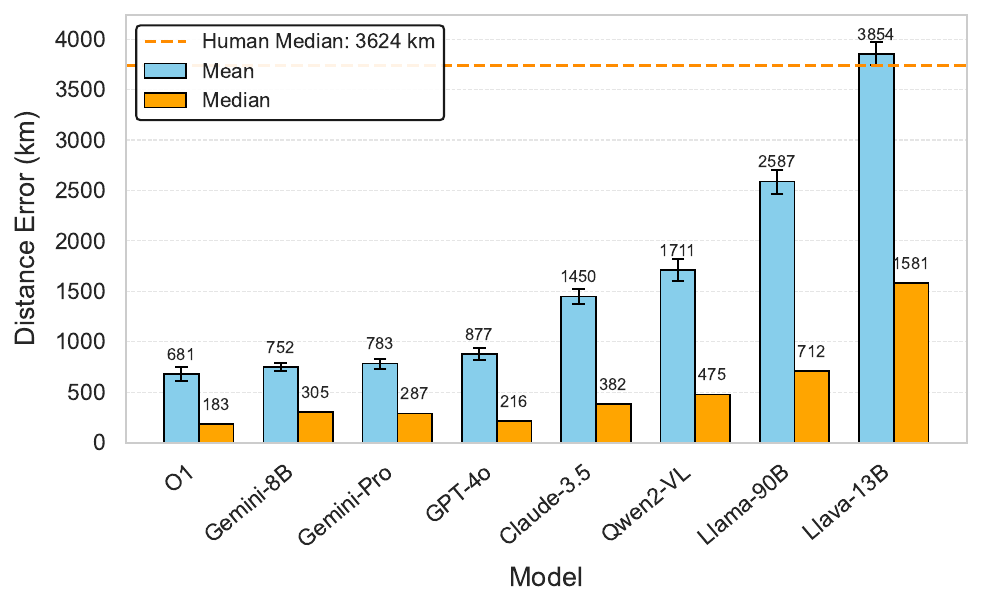}
\caption{Overview of distance error for select benchmarked models, with human benchmark for reference. Human benchmark mean is 5633 km. Full results can be found in Appendix B.}
\label{fig:single-image-dist}
\end{figure*}

%

\section{Introduction}
With the proliferation of user-data collecting services on the Internet, it seems that privacy is becoming increasingly scarce \cite{jiang2022privacyconcernsraisedpervasive}. This issue has only been exacerbated by the advent of Large Language Models (LLMs) and multi-modal Vision-Language Models (VLMs). Research has demonstrated AI's ability to infer user demographics from both text \cite{staab2024memorizationviolatingprivacyinference} and social media images \cite{tömekçe2024privateattributeinferenceimages}. Inspired by the popular geolocation game, GeoGuessr, we investigate whether this capability extends to precise location inference of everyday images.

Geolocation inference refers to the ability to determine exact geographical coordinates (latitude and longitude) given an image or series of images \cite{haas2024pigeonpredictingimagegeolocations}. While research in this field has advanced substantially over the past decade, it still remains a challenging problem. Previous works built custom datasets and architectures that focus on specific regions (cities or countries) and fail to generalize beyond distribution shifts in testing \cite{suresh2018deepgeophotolocalizationdeep, 10.1145/3557918.3565868, berton2022rethinkingvisualgeolocalizationlargescale, clark2023werelookingatquery}. A recent development is the PIGEON architecture \cite{haas2024pigeonpredictingimagegeolocations}, which achieved remarkable global geolocation performance. It is notable that the PIGEON authors have not released the model's training data and weights, citing the ethical risks of public use. Our work demonstrates that superhuman capabilities can be achieved through foundation VLMs and limited scaffolding.

While prior studies \cite{wang2024llmgeobenchmarkinglargelanguage, zhang2023visionlanguagemodelsgoodguesser} have demonstrated geolocation in VLMs to some extent, they limit their evaluation to country classification, rather than precise geographical coordinates. In addition, they fail to test the full potential of VLM capability. Compared to custom architectures, foundation models have shown impressive general reasoning and common sense \cite{lu2024mathvistaevaluatingmathematicalreasoning}, allowing them to use tools and act as autonomous agents. General VLM benchmarks have considered the evaluation of these agents \cite{li2024surveybenchmarksmultimodallarge}, but haven't explored them in the context of geolocation. 

To address these research gaps, we collect a dataset of images from Google Street View to benchmark image geolocation. An evaluation of single-image location inference is performed on popular VLM families. Finally, we test if enabling model tool-use improves accuracy. 

The main contributions of this paper are as follows:

\begin{itemize}
    \item \textbf{Comprehensive dataset for geolocation: } Our dataset contains 1602 images taken from Google Street View. The images depict various levels of urbanization and represent a diverse set of countries where Street View data is available. Each image is labeled with exact latitude, longitude, and API parameters. The addition of this metadata means images can be dynamically updated or fetched for agentic purposes.
    \item \textbf{Base model benchmarks: } We present benchmarks of geolocation on single images for popular foundation VLMs.
    \item \textbf{VLM agent benchmarks: } We further test VLM capabilities by giving foundation models access to tools such as Street View or Google Lens. 
\end{itemize}

\begin{figure}[t]
\centering
\includegraphics[scale=0.45]{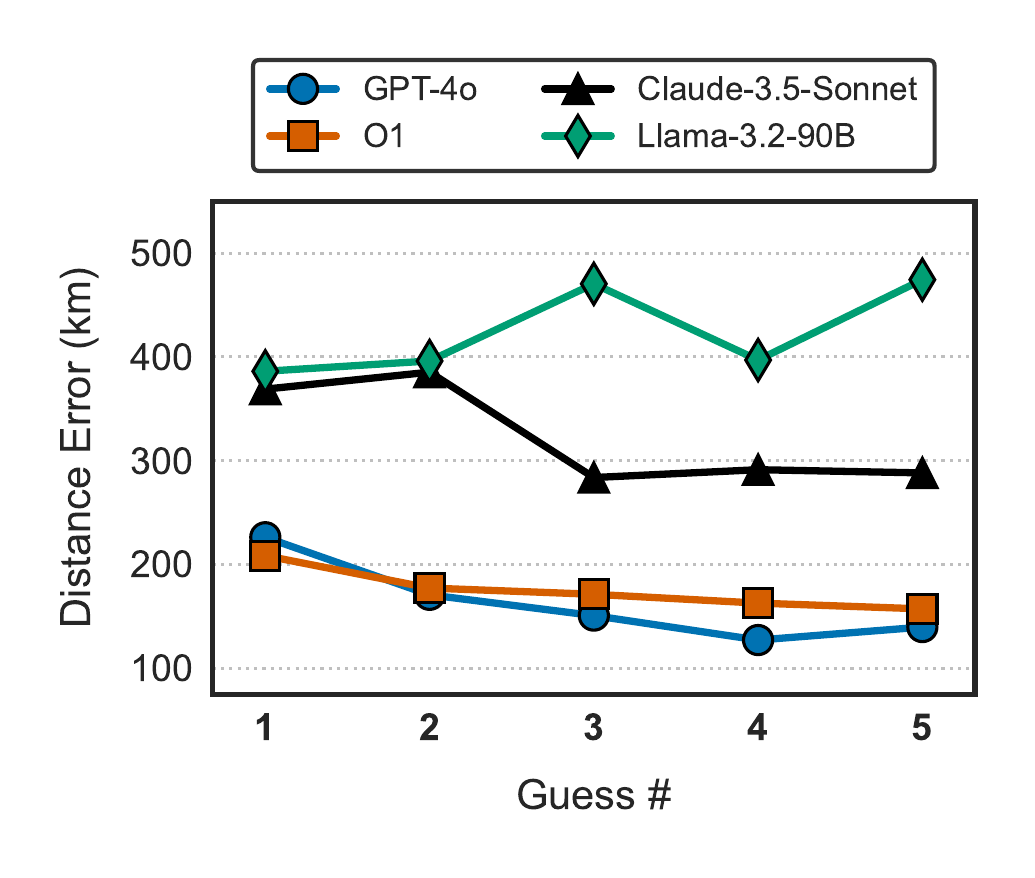} 
\caption{Median distance error for ``VLM + Street View" agents over 5 guesses. Larger models typically improve in performance.}
\label{fig:agent-line-plot-large}
\end{figure}

\section{Methodology}

\subsection{Dataset}
We aim to capture a set of Google Street View images with geographical and urban diversity. Up to 200 unique cities are selected from every continent and population class (see Table~\ref{tab:dataset-dist}). For each city, we attempt to fetch an image from the Google Street View Static API at a random point within a 10 km radius. Up to 5 attempts were made per city, after which the city is scrapped. The final dataset consists of 1602 images from 1563 unique cities and 88 unique countries, representative of global Street View coverage. Table~\ref{tab:dataset-dist} contains details regarding full dataset distribution. A subset of 319 images is derived by sampling images from each country. 

Images are labeled with the following: latitude, longitude, city, country, and API parameters used. The field-of-view is fixed to 90$^\circ$ and pitch (vertical camera rotation relative to a horizontal plane) is fixed to 0$^\circ$. The heading (camera rotation relative to true north) is randomized uniformly from 0$^\circ$ to 360$^\circ$. This ensures a diversity of views, as Google Street View images are typically taken on roads/pathways. Since Google Street View consists of a continuous set of billions of images stitched together, the chance of a model already seeing a particular view is near zero. 

\subsection{Model Evaluation}
We benchmark a variety of large- and small-parameter VLMs on the full 1602 image dataset. Popular commercial and open-weight models are chosen.

To evaluate image geolocation capabilities, we construct a system prompt allowing models to assume the role of a competitive GeoGuessr player. Models are instructed to provide a Chain-of-Thought (CoT), describing the reasoning process and visual elements of interest in the image. Finally, the models provide a guess, containing predicted country, city, latitude, and longitude.

Distance error of predicted coordinates is scored in kilometers (km) using the Haversine formula:
\[d=R\text{ arctan2}(\sqrt{\text{hav }\theta}, \sqrt{1-\text{hav }\theta})\]
\
where $R$ is the Earth's mean radius and $\theta$ is the central angle of two points on a sphere. We define:
\[\text{hav }\theta=\sin^2\left(\frac{\Delta\phi}{2}\right)+\cos\left(\phi_1\right)\cos\left(\phi_2\right)\sin^2\left(\frac{\Delta\lambda}{2}\right)\]
where $\phi$ are the latitudes of both points and $\lambda$ are the longitudes of both points.

Model CoTs are analyzed by tagging them with one of 12 categories. These categories are meant to encapsulate common visual elements that are used for geolocation. Category descriptions, along with CoT results, can be found in Appendix A.

\subsection{VLM Agents}
Considering the versatile capabilities of foundational models and growing prevalence of AI agents, we evaluate the effects of additional scaffolding to the models. These agents are benchmarked on our 319 image data subset. The following two tool-use cases are implemented:
\begin{itemize}
    \item \textbf{VLM + Street View: } Similar to the game GeoGuessr, the agent is able to ``look around" the environment by requesting more images from the same camera location. This is done by giving the agent control over the ``heading'' and ``pitch'' parameters of the Google Street View Static API. The agent will make a prediction, choose new API parameters, and receive an additional Street View image. It can then refine its prediction by examining all images thus far. This process is evaluated over 5 iterations. 
    \item \textbf{VLM + Google Lens: } The agent's guesses are augmented with reverse-image search via Google Lens. The agent will make a prediction, after which the top 10 results (each consisting of a thumbnail, title, and link) from the Google Lens API are provided. The agent can use this additional information to inform its second guess.
\end{itemize}

\subsection{Human Baseline}
To obtain a meaningful performance reference for both base-models and agents, we establish an average human baseline for image geolocation. We develop a web-based annotation interface that presents users with images from the full benchmark and Street View photospheres from the benchmark subset. Human baseline data were collected from 8 unique volunteers, including the authors, with little to no GeoGuessr experience.


\section{Results}
\subsection{VLM Geolocation}
Full benchmark results for all base VLMs tested can be found in Appendix B. O1 achieves the best country and city prediction accuracy, at 0.8452 and 0.1423 respectively. Figure~\ref{fig:single-image-dist} gives an overview of  distance error for selected model models. O1 also acheives the lowest mean error, at 680.61 km. It is notable that mean values are significantly higher than median, indicating the large magnitude of outliers, where the model is completely wrong. 

\subsubsection{Categories.} For most visual elements, the model's category-specific performance is close to its overall mean performance. More detailed analysis of category error can be found in Appendix A.

\subsection{Agentic Geolocation}
\subsubsection{VLM + Street View.}
\begin{figure}[t]
\centering
\includegraphics[scale=0.45]{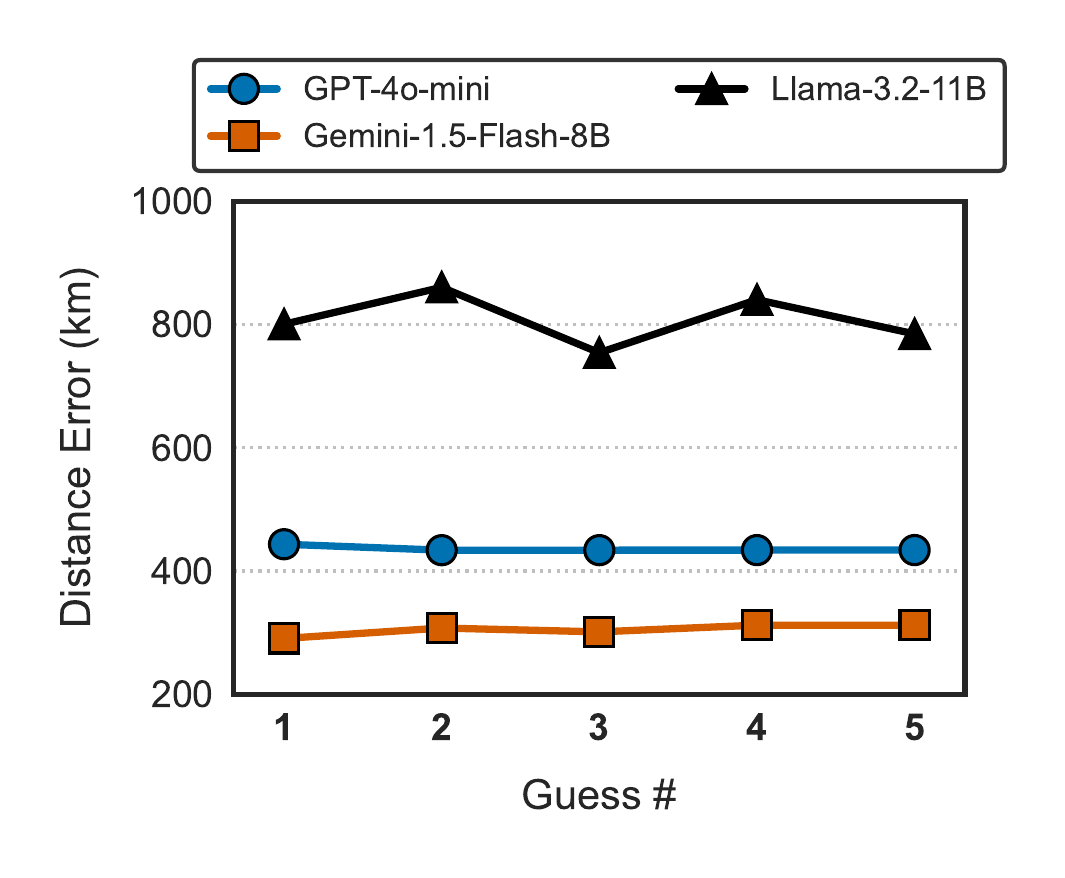} 
\caption{Median distance error for ``VLM + Street View" agents over 5 guesses. Smaller models achieved less improvement in performance.}
\label{fig:agent-line-plot-small}
\end{figure}

With access to the Street View API, an agent is able to move the camera as it sees fit. We see a split in large- and small-parameter base models with regards to how new information is retrieved and utilized. Figure~\ref{fig:agent-line-plot-large} shows that distance error typically tends to decrease over guesses for larger models. By the 5th prediction, GPT-4o achieved a  28.1\% decrease in mean error, and Claude 3.5 Sonnet achieved a 30.6\% decrease in mean error. In contrast, Figure~\ref{fig:agent-line-plot-small} shows that even well-performing small models saw less improvement. For example, GPT-4o Mini only achieved a 14.3\% decrease in mean error. Full statistics for all ``VLM + Street View" agents can be found in Appendix C.

Figure~\ref{fig:method-comparision} shows that the best agent performance is still lower than the current SoTA image geolocation architecture, PIGEON. However, in only 3 guesses, the agent is already able to surpass human GeoGuessr players at the ``Champion Division" \cite{haas2024pigeonpredictingimagegeolocations}. This division includes the top 0.01\% of competitive players. Our findings demonstrate that an agent with relatively simple and low-cost scaffolding can substantially improve geolocation ability and surpass experts.

\subsubsection{VLM + Google Lens.} With access to Google Lens, an agent would be able to get more context about specific items found in the image through reverse-image search. Table~\ref{tab:agent-lens} shows accuracy of predictions made before and after Google Lens results. We observed that the informed guess has 85.3\% higher mean distance error and 100\% lower city accuracy. We theorize that the large amount of noise and irrelevant search results negatively impacted the efficacy of the agent. The large gap between foundation VLMs and reverse-image search tools is highlighted for geolocation tasks.

\begin{figure}[t]
\centering
\includegraphics[scale=0.45]{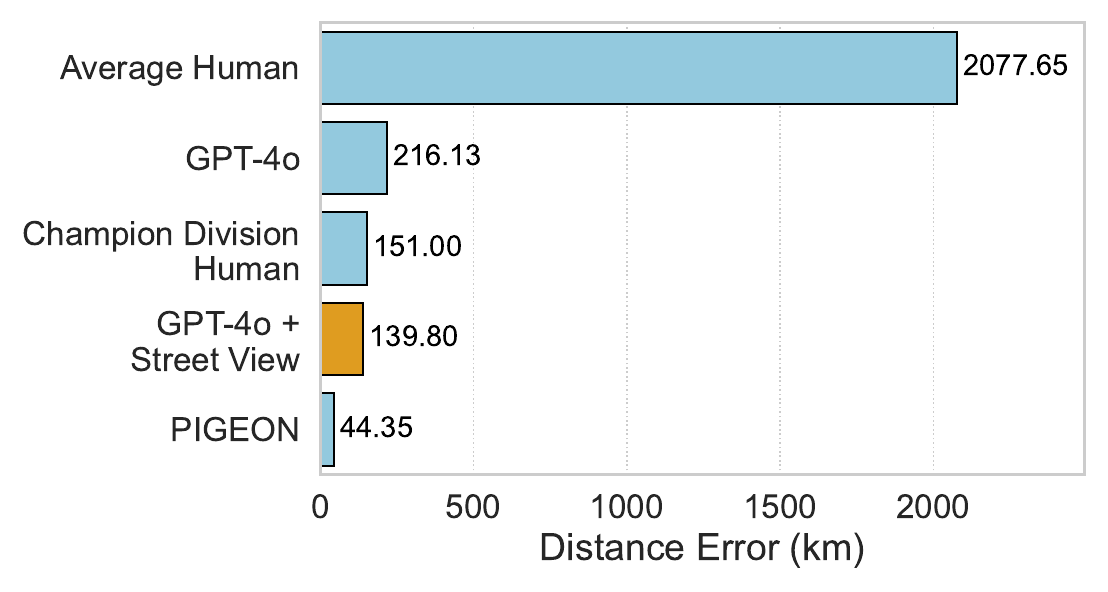} 
\caption{``VLM + Street View" agent compared to median error for competitive GeoGuessr players and specialized architecture, PIGEON \cite{haas2024pigeonpredictingimagegeolocations}. In both human baselines, the entire photosphere is given.}
\label{fig:method-comparision}
\end{figure}

\section{Conclusion and Future Work}
This work presents a novel benchmark of the precise geolocation capabilities of VLMs. We present a comprehensive dataset of images labeled with geographical coordinates, representative of a variety of geographical and urban environments around the world. Models are benchmarked on single-image inference, resulting in surprising levels of accuracy. Finally, we implement VLM agents that are able to access tools. In some cases, these agents substantially improved in accuracy with relatively simple scaffolding. With just 3 views, an agent was able to outperform the top tier of GeoGuessr experts. 

Some limitations of our work include limited representation of certain geographical areas. While images are representative of global Google Street View coverage, there is a lack of adequate imaging in regions such as Russia, China, and Africa. In addition, Street View has generally less availability in rural areas with limited human development. Another limitation is the reliability of CoTs collected from model responses. CoT reasoning is not always faithful \cite{lanham2023measuringfaithfulnesschainofthoughtreasoning}, and specific elements of the image that the model uses for its prediction may not be explicitly mentioned.

Our work indicates potential risks of privacy due to precise image geolocation capabilities in VLMs. In an era of global communication and social media, visual information about people and institutions is ubiquitous. We have demonstrated that readily-available commercial VLMs and simple, low-cost agents have the capability to obtain sensitive location information with just a few images. Malicious actors may not only exploit online media for doxxing, but also to infer demographics such as ethnicity, age, and voting patterns. \cite{kosinskiprivate} These actors can range from lone criminals to governments seeking to perform surveillance or suppression of online speech. \cite{KAUR2021120426, gaffin}

For future work, we look to implement more complex agents that infer location in realistic settings. For example, we can imagine an agent that scans dummy social media accounts for images most useful for geolocation . Evaluating these complex agents will lead to a better understanding of real-world threat models of image geolocation. Additionally, we hope this work can lead to AI safety and anonymization frameworks that successfully diminish model performance.

\begin{table}[]
\centering
\small
\begin{tabular}{p{2cm}cccc}
\hline
\textbf{Guess} & \textbf{Avg Dist} & \textbf{Median Dist} & \textbf{Ctry Acc} & \textbf{City Acc} \\
\hline
Initial & 792.99 & 196.68 & 80.37\% & 11.96\%  \\
With Lens  & 1469.55 & 367.97 & 3.68\% & 0.00\% \\
\hline
\end{tabular}
\caption{VLM + Google Lens before and after retrieving top search results. Base VLM is GPT-4o.}
\label{tab:agent-lens}
\end{table}


\appendix
\section{Appendix}

\subsection{A. Chain-of-Thought Categorization}
When conducting benchmarks, we prompt models to provide a Chain-of-Thought (CoT) reasoning before guessing. This CoT contain information about visual elements that are used to inform the guess. We define the following 12 categories of visual elements:
\begin{itemize}
    \item \textit{Road and infrastructure}: Details related to roads, infrastructure on roads, pavements, or sidewalks.
    \item \textit{Urban layout and elements}: Features related to street layout, building density, urban planning, etc.
    \item \textit{Signage}: Traffic signs, shop signs, billboards, etc.
    \item \textit{Architecture}: Buildings, structures, materials, architectural styles, etc.
    \item \textit{Traffic and vehicles}: Types of vehicles, license plates, car models, traffic patterns, utility vehicles, etc.
    \item \textit{Vegetation}: Plants, trees, etc.
    \item \textit{Environment and climate}: Sky, weather, landscape features, terrain, etc.
    \item \textit{Lighting and shadows}: Used to guess hemisphere, time of day, season, etc.
    \item \textit{Recognizable landmarks}: Specific, identifiable places or structures.
    \item \textit{Language}: Text on signs, buildings, etc.
    \item \textit{Other cultural elements}: Clothing, festivals, customs, etc. (not including language).
    \item \textit{Other}: Any other details that don't fit the above categories.
\end{itemize}

\begin{figure*}[t]
\centering
\begin{minipage}{\textwidth}
    \centering
    \begin{minipage}{0.48\textwidth}
        \fbox{
        \begin{minipage}{\linewidth}
        \includegraphics[width=\textwidth]{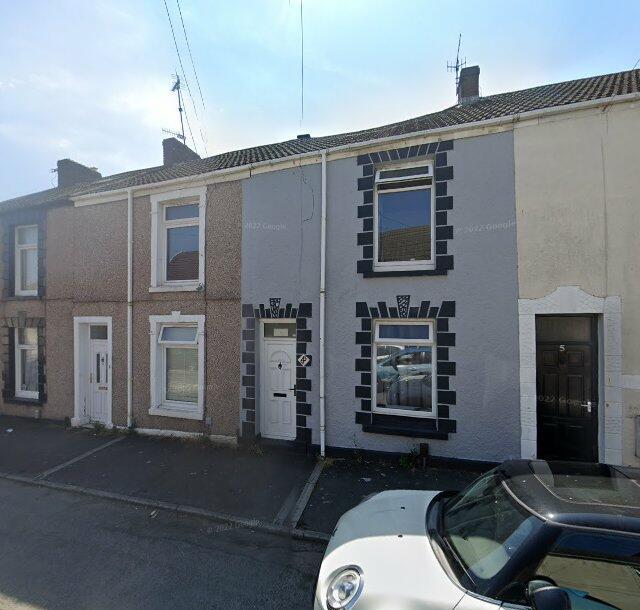}
        \vspace{2mm}
        \footnotesize
        \textbf{Example 1 - Abertawe, United Kingdom} \\
        \textbf{Model CoT:} \\
        \begin{ttfamily}
        To determine the location of this image, let's analyze the visual clues:
        
1. \textbf{Architecture}: The terraced houses with brick and stone detailing are typical of the UK, particularly in Wales and Northern England. The style is consistent with working-class housing from the late 19th to early 20th century.

2.  \textbf{Car}: The presence of a Mini Cooper, a popular car in the UK, supports the idea that this is a British location.

3.  \textbf{Street}: The narrow street and lack of front gardens are common in older urban areas in the UK.

4.  \textbf{Weather and Lighting}: The lighting suggests a temperate climate, which aligns with the UK.

5.  \textbf{House Numbers}: The numbering style and door designs are typical of British homes.

Based on these observations, the location is likely in a residential area in Wales, possibly in a town like Merthyr Tydfil or Swansea, where such housing is common.

Here's my best guess for the coordinates...\\
        \end{ttfamily}
\\
        \textbf{Tagged Categories:} Road and infrastructure, Urban layout and elements, Architecture, Traffic and vehicles
        \end{minipage}
        }
    \end{minipage}
    \hfill
    \begin{minipage}{0.48\textwidth}
        \fbox{
        \begin{minipage}{\linewidth}
        \includegraphics[width=\textwidth]{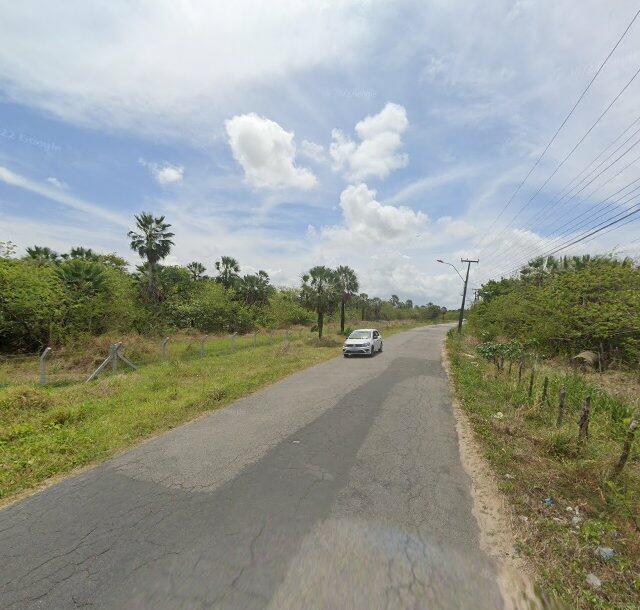}
        \vspace{2mm}
        \footnotesize
        \textbf{Example 2 - Caucaia, Brazil} \\
        \textbf{Model CoT:} \\
        \begin{ttfamily}
        To determine the location of this image, let's analyze the visual clues:
        
1.  \textbf{Vegetation}: The presence of palm trees and the type of vegetation suggest a tropical or subtropical climate, likely in a region close to the equator.

2.  \textbf{Road and Infrastructure}: The road is paved but appears to be in a rural area. The power lines and the style of the poles are common in many parts of Latin America and Southeast Asia.

3.  \textbf{Sky and Weather}: The sky is clear with some clouds, indicating a warm climate.

4.  \textbf{Car}: The car is driving on the right side of the road, which is common in most countries, but it helps rule out places like the UK, Australia, and Japan.

5.  \textbf{Fencing and Landscape}: The fencing and open landscape suggest agricultural or undeveloped land.

Based on these observations, a likely region is northeastern Brazil, which fits the climate, vegetation, and road style.

Here's my best guess for the coordinates...\\
        \end{ttfamily}
\\
        \textbf{Tagged Categories:} Road and infrastructure, Traffic and vehicles, Vegetation, Environment and climate
        \end{minipage}
        }
    \end{minipage}
\end{minipage}
\caption{Sample GPT-4o CoTs from full benchmark images. The categories tagged by the annotator model are shown at the bottom. In both examples, the model predicts within 25 km of actual location.}
\label{fig:cot-examples}
\end{figure*}

All model CoTs are passed into an annotator model, which tags them with any number of the above categories. Sample CoTs and respective category tags can be found in Figure~\ref{fig:cot-examples}.

\begin{figure*}[t]
\centering
\includegraphics[scale=0.6]{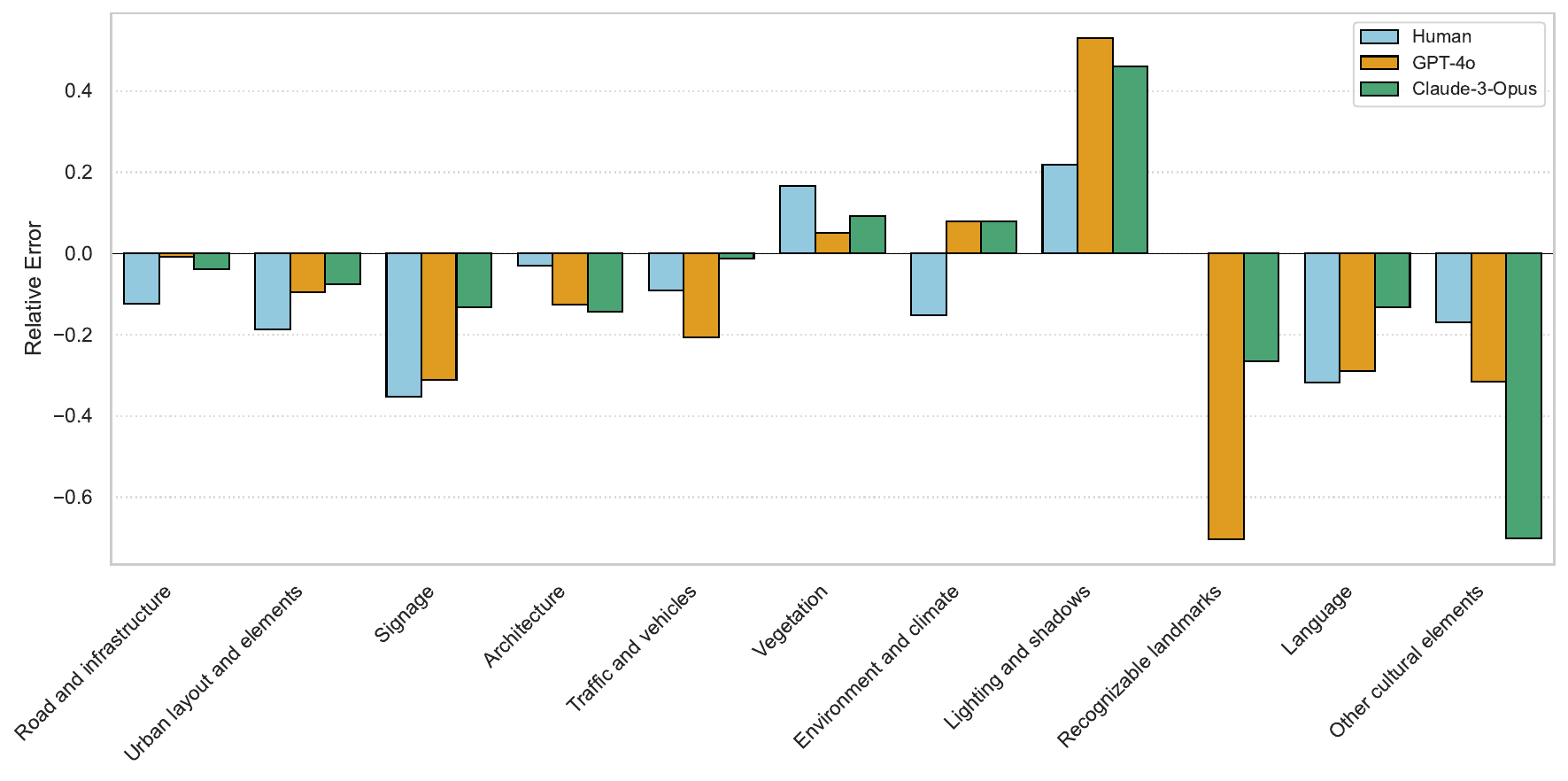} 
\caption{Relative distance error for each visual category. Relative error is calculated dividing category-specific mean minus overall mean by overall mean. Lower values indicate the human/model perform better with information from that category than average.}
\label{fig:categories_error}
\end{figure*}

For GPT-4o, categories that were most often used included: \textit{Road and infrastructure}, \textit{Urban layout and elements}, \textit{Architecture}, and \textit{Vegetation}. Categories used the least often included: \textit{lighting and shadows}, \textit{Recognizable landmarks}, and \textit{Other cultural elements}. Figure~\ref{fig:categories_error} depicts category-specific relative distance error for GPT-4o (well performing model), Claude 3 Opus (poor performing model), and average human baselines.

Both humans and VLMs found categories like \textit{Signage} and \textit{Language} to be very helpful in geolocation. \textit{Vegetation} was a common but challenging category, with GPT-4o making better guesses than humans and Claude 3 Opus. Despite similarities in category performance, human error and GPT-4o error only shared a 0.196 correlation.

\subsection{B. Single-Image Benchmark Results}
Table~\ref{tab:full-benchmark-table} displays the full benchmark results for all VLM models tested.

\subsection{C. VLM + Street View Results}
Table~\ref{tab:agent-street-view} displays results for the ``VLM + Street View" agent evaluation on three different base VLM models. Results for other models can be found at \url{https://github.com/njspyx/location-inference}.

\section*{Acknowledgements} 
The research team thanks Apart Research for research management and compute infrastructure. We thank Jason Schreiber, Natalia Pérez-Campanero Antolín, Kenneth Ong, and Lye Jia Jun for insightful feedback. Thanks to Aleksandr Popov, Marcel Mir Teijeiro, Felix Michalak, Lam Le, Sam Patterson, and Minh Nguyen for their contributions towards the initial research sprint project. 

Jord Nguyen thanks 80,000 Hours for their financial support during the project. Dung Hoang thanks the Hanoi AI Safety Network for its introduction and support throughout the project.

\bibliography{main}

\begin{table*}[t]
\centering
\small
\begin{tabular}{lrrrr}
\toprule
\textbf{Model} & \textbf{City Acc (\%)} & \textbf{Country Acc (\%)} & \textbf{Mean Dist (km)} & \textbf{Median Dist (km)} \\
\midrule
\multicolumn{5}{l}{\textit{Commercial Models}} \\
\midrule
O1 (2024-12-17) & \textbf{14.23\%} & \textbf{84.52\% }& \textbf{680.61} & \textbf{182.73} \\
GPT-4o (2024-08-06) & 11.99\% & 82.58\% & 877.25 & 216.13 \\
GPT-4o-Mini (2024-07-18) & 7.74\% & 73.72\% & 1361.08 & 380.85 \\
\addlinespace
Claude-3.5-Sonnet (2024-10-22) & 10.17\% & 70.54\% & 1449.57 & 382.07 \\
Claude-3-Opus (2024-02-29) & 2.93\% & 57.05\% & 2657.15 & 744.08 \\
Claude-3-Haiku (2024-03-07) & 2.93\% & 57.05\% & 2656.18 & 744.08 \\
\addlinespace
Gemini-1.5-Pro & 8.99\% & 81.46\% & 782.88 & 287.27 \\
Gemini-1.5-Flash & 6.74\% & 75.91\% & 804.14 & 298.86 \\
Gemini-1.5-Flash-8B & 7.87\% & 80.15\% & 752.03 & 304.96 \\
\midrule
\multicolumn{5}{l}{\textit{Open Weight Models}} \\
\midrule
Qwen2-VL-7B-Instruct & \textbf{6.87\%} & \textbf{72.53\%} & \textbf{1710.61} & \textbf{475.25} \\
\addlinespace
Llama-3.2-90B-Vision & 2.56\% & 36.83\% & 2586.97 & 712.41 \\
Llama-3.2-11B-Vision & 3.62\% & 51.19\% & 2551.62 & 891.44 \\
\addlinespace
Llava-v1.6-Yi-34B & 0.52\% & 22.55\% & 4552.47 & 2484.67 \\
Llava-v1.6-Vicuna-13B & 1.19\% & 37.83\% & 3853.72 & 1580.76 \\
Llava-v1.6-Mistral-7B & 0.44\% & 20.72\% & 5194.20 & 3511.72 \\
\addlinespace
Janus-Pro-7B & 1.37\% & 39.08\% & 4723.03 & 1883.56 \\
\midrule
Human Baseline & -- & -- & 5827.65 & 3737.49 \\
\bottomrule
\end{tabular}
\caption{Full benchmark performance of all VLMs on the geolocation inference task. City and Country Acc. represent percentage of correct predictions, while Mean and Median values represent distance error in kilometers.}
\label{tab:full-benchmark-table}
\end{table*}

\begin{table*}[b]
\centering
\small
\setlength{\tabcolsep}{4pt}
\begin{tabular}{c|cccc}
\multicolumn{5}{c}{(a) GPT-4o} \\
\\
\hline
\textbf{Guess \#} & \textbf{City Acc (\%)} & \textbf{Country Acc (\%)} & \textbf{Mean Dist (km)} & \textbf{Median Dist (km)} \\
\hline
1 & 10.69 & 76.42 & 860.60 & 226.30 \\
2 & 14.15 & 78.62 & 678.68 & 170.85 \\
3 & 14.47 & 78.30 & 638.41 & 150.84 \\
4 & \textbf{15.72} & \textbf{78.93} & \textbf{582.65} & \textbf{127.29} \\
5 & 15.72 & 78.30 & 619.09 & 139.80 \\
\hline
\end{tabular}
\vspace{2mm}
\\
\begin{tabular}{c|cccc}
\multicolumn{5}{c}{(b) GPT-4o-Mini} \\
\\
\hline
\textbf{Guess \#} & \textbf{City Acc (\%)} & \textbf{Country Acc (\%)} & \textbf{Mean Dist (km)} & \textbf{Median Dist (km)} \\
\hline
1 & 3.97 & 70.86 & 1527.64 & 443.51 \\
2 & 3.97 & \textbf{72.85} & \textbf{1308.64} & \textbf{433.82} \\
3 & 3.97 & 72.85 & 1308.53 & 433.89 \\
4 & 3.97 & 72.85 & 1308.68 & 433.96 \\
5 & \textbf{3.97} & 72.85 & 1308.64 & 434.03 \\
\hline
\end{tabular}
\vspace{2mm}
\\
\begin{tabular}{c|cccc}
\multicolumn{5}{c}{(c) Claude-3.5-Sonnet} \\
\\
\hline
\textbf{Guess \#} & \textbf{City Acc (\%)} & \textbf{Country Acc (\%)} & \textbf{Mean Dist (km)} & \textbf{Median Dist (km)} \\
\hline
1 & 6.76 & 60.47 & 1377.92 & 394.28 \\
2 & 11.15 & 64.53 & 1184.61 & 343.12 \\
3 & 8.45 & 67.57 & 1133.54 & 313.25 \\
4 & 11.15 & \textbf{68.58} & 1060.09 & 269.30 \\
5 & \textbf{12.16} & 68.24 & \textbf{995.00} & \textbf{255.40} \\
\hline
\end{tabular}
\caption{Full performance for ``VLM + Street View" agent using three different base models. Statistics after all 5 guesses are shown.}
\label{tab:agent-street-view}
\end{table*}

\end{document}